\documentclass[letterpaper, 10 pt, conference]{ieeeconf}  
\usepackage{cite}
\usepackage[ruled,vlined]{algorithm2e}
\usepackage{graphicx}
\usepackage{float}
\usepackage{url}
\usepackage[hidelinks]{hyperref}
\usepackage{amsmath,amssymb,amsfonts}
\usepackage{textcomp}
\usepackage{booktabs}
\usepackage[table]{xcolor}
\usepackage{placeins}

\IEEEoverridecommandlockouts                              

\title{\LARGE \bf

BO-COMM: Bidirectional Optical System for Simultaneous Localization and Communication in Swarm Robotics

}

\author{$^\dag$Cheng Hu$^{1,3*}$, ${^\dag}$Haotian Wu$^{2*}$, ${^\dag}$Qi Lin$^{2}$,  Manrui Zheng$^{1}$, Shigang Yue$^{2}$, Tao Zou$^{1}$ and Jigen Peng$^{3,4}$
\thanks{$^\dag$\ These authors contributed equally to this work.}%
\thanks{$^{1}$The authors are with the School of Mechanical and Electrical Engineering, Guangzhou University, Guangzhou 510006, China.
        }%
\thanks{$^{2}$The authors are with the School of Computing and Mathematical Sciences, University of Leicester, Leicester, United Kingdom, LE1 7RH.}%
\thanks{$^{3}$The authors are with the Machine Life and Intelligence Research Centre,
        Guangzhou University, Guangzhou, China, 510006.
        }%
\thanks{$^{4}$The authors are with the School of Mathematics and Information Science, Guangzhou University, Guangzhou 510006, China.
        }%
\thanks{%
  \parbox[t]{\dimexpr\columnwidth-1em\relax}{%
    \raggedright
    \textsuperscript{*}Corresponding authors:\protect\newline
    Cheng Hu: \texttt{c\_hu@gzhu.edu.cn}\protect\newline
    Haotian Wu: \texttt{hw386@le.ac.uk}%
  }%
}%
}

\begin{document}

\maketitle
\thispagestyle{empty}
\pagestyle{empty}

\begin{abstract}
Swarm robotics experiments need to track where robots are, know what state each robot is in, and feed environmental information back to them. Reporting states over a shared radio channel can suffer from contention as the number of robots grows, while static fiducial markers provide pose but cannot report changing states. We present BO-COMM, a bidirectional optical system that observes pose and state through one camera and returns environmental cues through light. Each Colias IV robot, fitted with a CEU-PLUS extension unit, carries a ring of 12 LEDs: a five-bit marker encodes heading, and the remaining seven bits display one of 52 state codes. An overhead camera locates robot centers by ring-based voting and decodes position, heading and state for all visible robots from a single frame. In the downstream direction, a projector renders spatial cues such as virtual pheromone fields, and an infrared channel delivers addressed commands. At $2048\times1280$ resolution, image-processing time increases only modestly from 2 to 30 robots. Path-following, color-reporting and multi-pheromone experiments show the system closing the loop between optical state reports and projected environmental feedback in a $2.50\times1.56$~m arena.
\end{abstract}

\section{Introduction}\label{sec:introduction}
Swarm robotics investigates collective behavior arising from interactions among multiple robots\cite{swarm1,swarm2}. Physical experiments require both observation and communication: robot positions and headings must be obtained, changing operational states may need to be reported, and environmental cues or commands must reach the robots. These functions can impose different requirements on the experimental infrastructure.

Fiducial markers such as AprilTag\cite{apriltag} encode identifiers in fixed visual patterns and support camera-based pose estimation. A static marker, however, does not itself report a changing task state or sensor value. Such information requires an additional mechanism. Radio-frequency (RF) communication is one option, but shared-channel wireless systems can face contention and congestion as concurrent communication demands increase. The severity depends on the protocol, traffic load, and deployment conditions. More generally, communication conflicts can constrain swarm operation, as illustrated by studies of congestion in collective decision-making\cite{congestion}. These considerations motivate alternative ways to collect selected state reports alongside pose observations.

\begin{figure}[t]
    \centering
    \includegraphics[width=\linewidth]{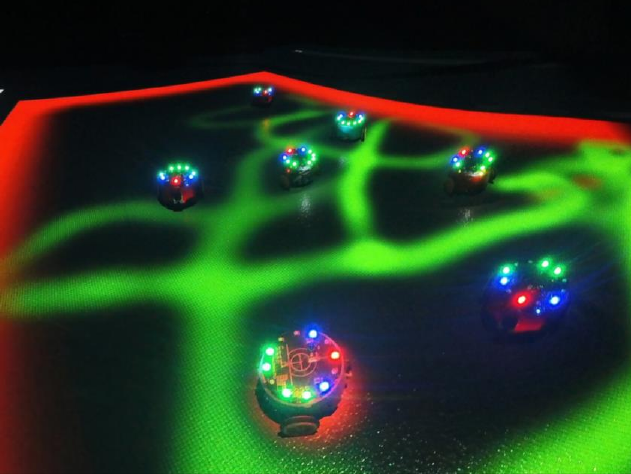}
    \caption{BO-COMM in operation. Robot-mounted LED rings display spatial patterns that carry state information and provide geometric cues for pose estimation. An overhead camera observes the rings. Projected light supplies environmental cues sensed by the robots.}
    \label{fig1}
\end{figure}

BO-COMM combines these functions through optical channels (Fig.~\ref{fig1}). An overhead camera captures the robot-mounted LED rings, from which the processor estimates position, orientation, and a dynamic code. This optical camera communication (OCC) pathway provides an alternative to RF transmission for selected upstream state reports. When robot signals are spatially distinguishable in the image, multiple reports can be acquired in one frame, subject to visibility, image resolution, and background illumination.

The downstream pathway combines projected visible light with an infrared command channel. Projection represents environmental information spatially, for example as virtual pheromone fields inspired by chemical communication\cite{pheromones}, whereas infrared transmission carries addressed commands. The architecture therefore reflects the asymmetric requirements of the experiments considered here: many robots report compact states to one observer, while the environment distributes spatial cues and the command channel reaches designated receivers.

The contribution is the integration of three elements: joint pose and state decoding from an LED ring; optical upstream observation with projected environmental feedback and infrared command reception; and functional demonstrations on a Colias IV/CEU-PLUS implementation. The evaluation concerns a planar arena and the hardware described below.

Section~\ref{sec:related} places this design in context. Sections~\ref{sec:pipeline} and~\ref{sec:methods} describe the architecture and processing methods. Section~\ref{sec:experiments} presents the demonstrations, and Section~\ref{sec:discussion} discusses their scope and limitations.

\section{Related Work}\label{sec:related}
\subsection{Indoor Robot Localization}
Indoor localization methods include fiducial markers\cite{apriltag,ArTag,whycode,accurate}, motion capture\cite{MotionCapture1,MotionCapture2}, visual simultaneous localization and mapping\cite{SLAM}, and RFID-based methods\cite{RFID}. These approaches differ in sensing geometry, onboard requirements, calibration, and infrastructure. Fiducial markers and motion capture are relevant to externally observed robot experiments, whereas SLAM and RFID address other combinations of sensing and positioning requirements.

The distinction emphasized here is between a fixed identifier and a visual pattern that can change with the robot's state. BO-COMM retains a finite code set and depends on camera visibility. Its LED pattern supports state reporting as well as pose estimation, but this functionality alone does not imply better localization accuracy or greater identification capacity than existing marker systems.

\subsection{Optical Communication and Virtual Environments}
Optical communication includes Li-Fi\cite{LiFi}, color-based robot signaling\cite{colorcomm}, OCC\cite{OCC}, and camera-based localization using optical signals\cite{slac1}. Existing swarm platforms also combine virtual pheromones, visual environments, localization, and data transmission\cite{colcosphi}. Mixed-reality environments provide a broader context for connecting physical robot actions to programmable virtual information\cite{mixedreality}.

BO-COMM focuses on a particular combination: a spatially coded LED ring is used for upstream pose and state observation, projected light supplies local environmental cues, and infrared carries commands. The contribution lies in the hardware and processing integration described here. Spatially separated robot images provide a form of spatial multiplexing related to space division multiple access\cite{sdma}; in this implementation, usable separation is constrained by the camera image and the visibility of each ring.

\section{System Pipeline}\label{sec:pipeline}
\begin{figure}[t]
    \centering
    \includegraphics[width=\linewidth]{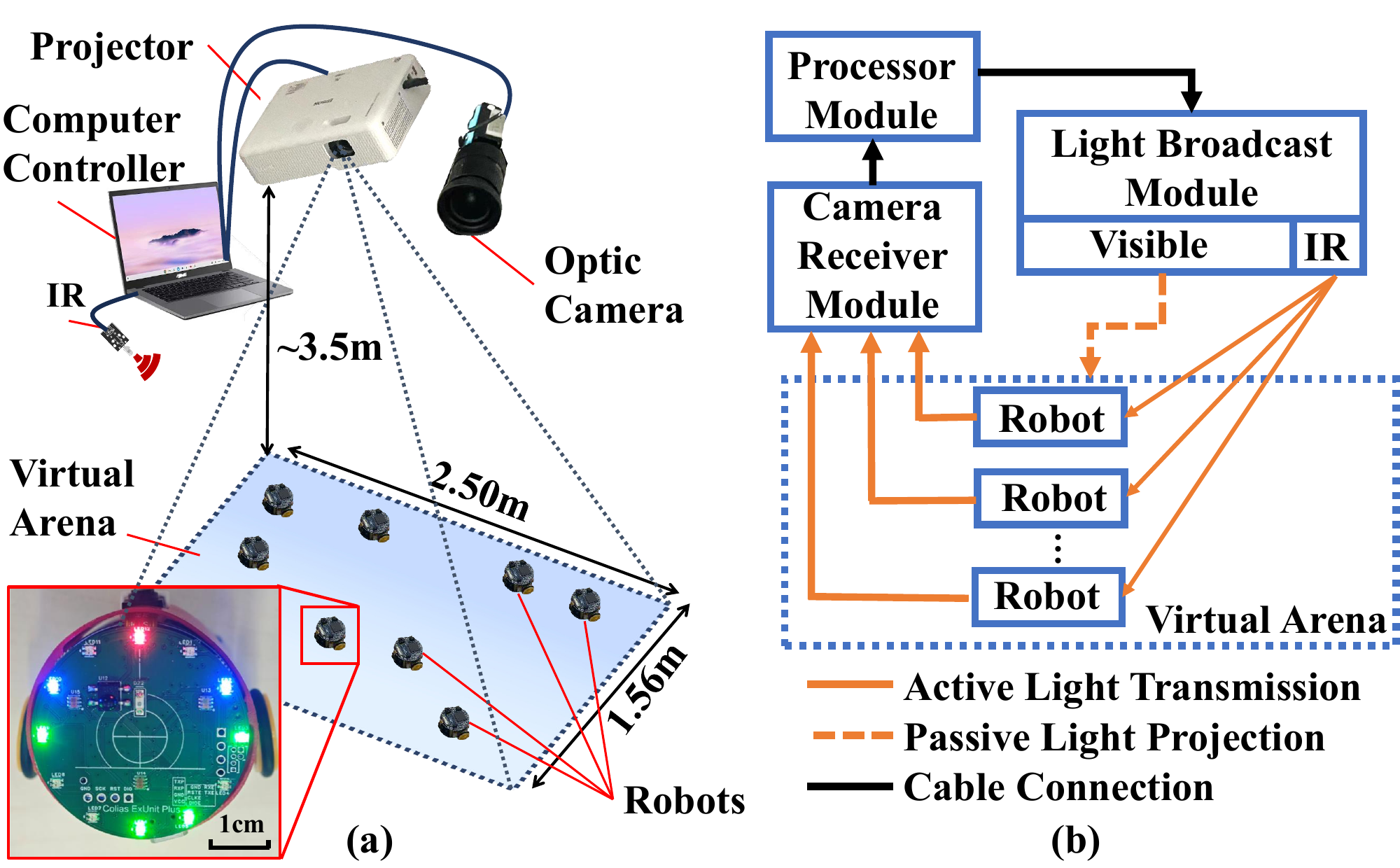}
    \caption{System architecture and data flow. (a) Computer, projector, overhead camera, infrared transmitter, and robots; robot hardware is described in Section~\ref{sec:hardware}. (b) Wired connections link the external modules to the computer. Robot LEDs provide upstream optical signals, while projected light and infrared provide downstream cues and commands.}
    \label{fig2}
\end{figure}

Figure~\ref{fig2} summarizes the processor, camera, light-broadcast modules, and robots. Upstream sensing uses actively emitted LED patterns. The camera records their geometry and on/off states, allowing the processor to estimate robot pose and decode state information from the same image.

Each robot carries 12 LEDs arranged in a ring. The coding design shown in Fig.~\ref{fig5} contains 52 available information codes after excluding patterns that interfere with the orientation marker or contain long unlit sequences. These are code values, not necessarily permanent robot identities. Their meaning depends on the task, for example a sensed color or a behavioral state. Robots may display the same state code at different positions. Associating observations with persistent identities across frames is a separate problem from decoding the instantaneous pattern.

The visible downstream pathway uses projected patterns to represent local environmental cues. The infrared pathway\cite{infrared} carries command payloads and recipient identifiers. These pathways serve different functions: sensing projected illumination need not involve decoding an addressed digital packet. Together, they allow optical observations to inform changes to the environment, which in turn influence robot behavior.

\section{Hardware and Methods}\label{sec:methods}
\subsection{Hardware}\label{sec:hardware}
\begin{figure}[t]
    \centering
    \includegraphics[width=\linewidth]{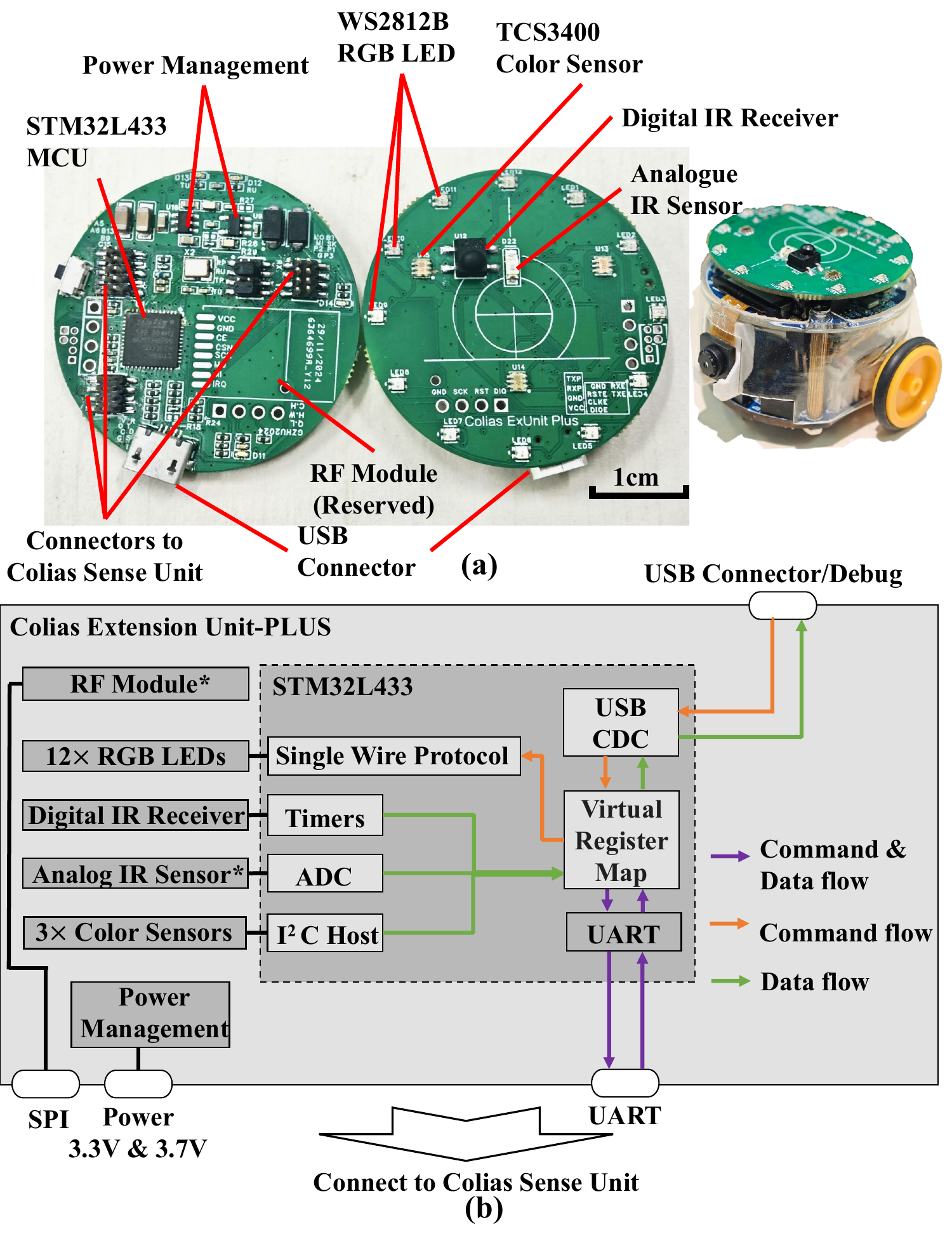}
    \caption{CEU-PLUS hardware. (a) Board layout and assembled robot. The 12 LEDs are spaced at $30^\circ$ intervals, with optical sensing components mounted on the upper surface. (b) Module connections and internal data flow. The analog infrared sensor is reserved, and the RF module provides backward compatibility.}
    \label{fig3}
\end{figure}

\subsubsection{Robot Extension Unit}
The implementation uses Colias IV robots\cite{colias} fitted with the Colias Extension Unit-PLUS (CEU-PLUS), shown in Fig.~\ref{fig3}. The extension connects optical sensing and signaling to the existing robot electronics. Twelve WS2812B LEDs form a circular array with $30^\circ$ angular spacing. Their arrangement provides the geometric basis for center estimation and spatial coding.

Three TCS3400 color sensors, each with red, green, blue, and clear channels, sense environmental illumination. A digital infrared receiver accepts downstream commands. The board also includes an analog infrared sensor that is reserved rather than used as an evaluated channel here. An STM32L433 microcontroller operating at 80~MHz manages the virtual register map and separates command and data flows. CEU-PLUS interfaces with the Colias Sense Unit; the RF module shown in the schematic supports backward compatibility.

\subsubsection{External Modules}
A Work5Power WP-UT880 camera with a native resolution of $4096\times2160$ observes an arena of approximately $2.50\times1.56$~m. Unless otherwise stated, the experiments use $2048\times1280$ images, corresponding to approximately 1.22~mm per pixel. An Epson CO-W01 projector renders environmental cues, and a digital infrared transmitter sends commands. An Intel Core i7 computer with a 2.80~GHz CPU performs image processing and environmental updates. This hardware configuration defines the implementation evaluated in this paper.

\subsection{Upstream Pose and State Decoding}
\subsubsection{Ring-Based Center Estimation}
The task is to estimate ring centers from visible LED spots and then read the spatial pattern around each center. Let the detected spot coordinates in an image be $\mathcal{X}=\{\mathbf{x}_i\}_{i=1}^{N}$, where $\mathbf{x}_i=(u_i,v_i)^\top$ is expressed in pixels. The method assumes approximately circular ring projections with a known image-space radius $r$. It also uses a specified robot count $n$ to limit the number of candidate centers. These assumptions concern the calibrated planar setup; they do not provide depth estimation or invariance to arbitrary viewing geometry.

Represent the detected spots by an impulse map
\begin{equation}
 S(\mathbf{u})=\sum_{i=1}^{N}\delta(\mathbf{u}-\mathbf{x}_i),
 \label{eq:spots}
\end{equation}
where $\delta$ denotes a unit impulse on the image grid. Define an unnormalized ring response kernel
\begin{equation}
 K(\mathbf{d})=\exp\!\left[-\frac{(\|\mathbf{d}\|_2-r)^2}{2\sigma^2}\right],
 \label{eq:kernel}
\end{equation}
where $r$ and $\sigma$ are in pixels. The width $\sigma$ controls tolerance to radial deviations and is distinct from the observed LED spot diameter. The kernel assigns geometric weights rather than probabilities.

Each spot contributes a ring-shaped response to possible center locations. The accumulated voting map is
\begin{equation}
 H(\mathbf{c})=(S*K)(\mathbf{c})
 =\sum_{i=1}^{N}K(\mathbf{c}-\mathbf{x}_i).
 \label{eq:voting}
\end{equation}
Here $*$ denotes spatial convolution. Spots lying near a common ring reinforce the response near its center. Candidate centers are obtained from spatially separated local maxima:
\begin{equation}
 \widehat{\mathcal{C}}=
 \operatorname{PeakSelect}_{n}\!\left(H;d_{\min}\right).
 \label{eq:centers}
\end{equation}
The operator retains up to $n$ local peaks in descending response order subject to a minimum separation $d_{\min}$ in pixels. The separation constraint expresses the physical spacing of robot centers. Missing spots, reflections, overlapping image features, and radius mismatch can affect the response, so the selected peaks are candidate centers rather than guaranteed detections.

\begin{figure}[t]
    \centering
    \includegraphics[width=\linewidth]{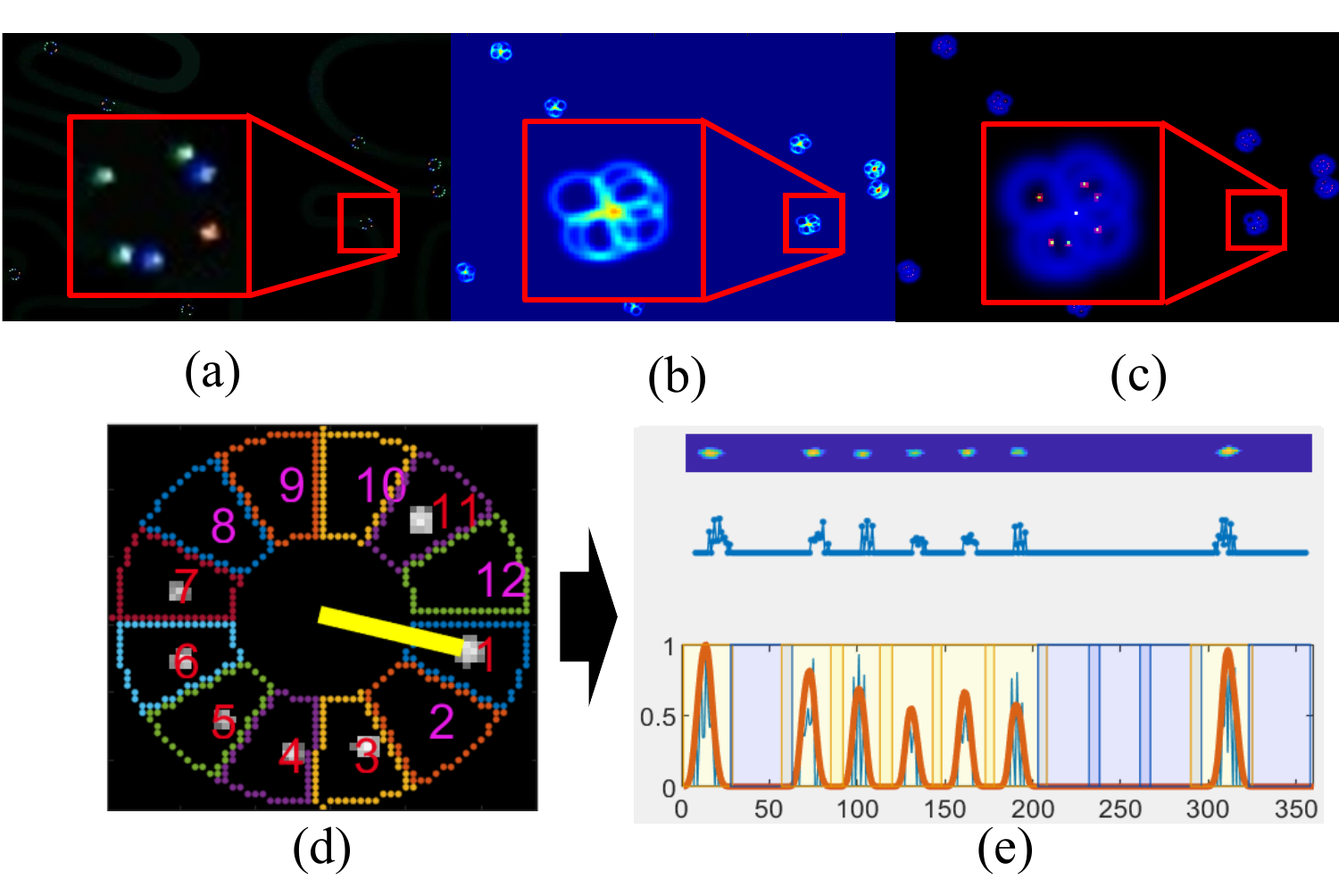}
    \caption{Upstream processing. (a) Input image and detected light spots. (b) Ring-based voting response. (c) Estimated centers overlaid on the spots. (d) An annular region divided into 12 numbered angular sectors; the yellow line indicates the estimated heading. (e) Unwrapped annulus, angular intensity profile, and sector decisions for reading the LED pattern.}
    \label{fig4}
\end{figure}

\subsubsection{Spatial Code and Orientation}
For an estimated center $\widehat{\mathbf{c}}_j$, the surrounding annular image region is unwrapped into an angular representation (Fig.~\ref{fig4}). Its 12 LED positions are spaced by $30^\circ$. Let $\mathcal{S}_{jk}$ denote the annular sector associated with position $k$ after angular alignment to the observed ring. The binary sequence is written
\begin{equation}
 \mathbf{b}_j=(b_{j0},\ldots,b_{j,11}),\qquad b_{jk}\in\{0,1\}.
 \label{eq:code}
\end{equation}
In the spot representation, the occupancy interpretation is
\begin{equation}
 b_{jk}=\mathbf{1}\{\mathcal{X}\cap\mathcal{S}_{jk}\ne\varnothing\}.
 \label{eq:occupancy}
\end{equation}
The indicator is one when a detected spot occupies the sector. In image-based decoding, the corresponding decision is obtained from the unwrapped angular intensity profile. Sector assignment presumes that the angular phase of the LED lattice has been identified; fixed image-axis sectors alone are insufficient for a rotated ring. Detection and intensity decisions therefore affect the recovered pattern.

\begin{figure}[t]
    \centering
    \includegraphics[width=\linewidth,trim=14 30 48 0,clip]{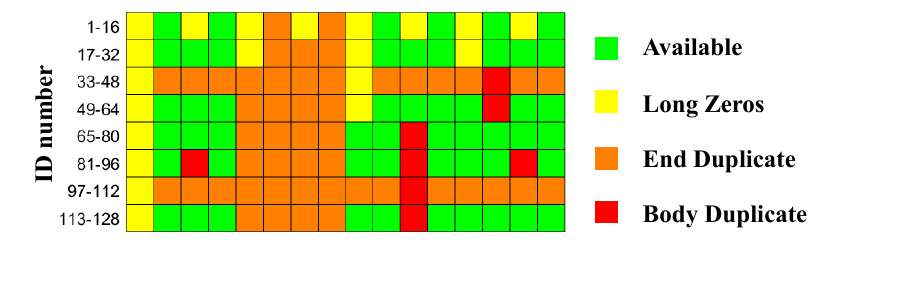}
    \caption{Codebook of the seven-bit information codes that follow the orientation marker 10101. In row $i$ ($1\le i\le8$) and column $j$ ($1\le j\le16$), the code consists of the four-bit binary form of $j-1$ followed by the three-bit binary form of $i-1$; row labels give the cell numbers. Codes are excluded when the ring contains four or more consecutive unlit LEDs (Long Zeros), or when the marker pattern also appears across the marker--code boundary (End Duplicate) or within the code (Body Duplicate). The remaining 52 codes are available.}
    \label{fig5}
\end{figure}

The orientation marker is $\mathbf{o}=(1,0,1,0,1)$ (Fig.~\ref{fig5}). Since an unwrapped sequence can start at any angular position, marker matching is circular. For $m\in\{0,\ldots,11\}$, the candidate marker starts are
\begin{equation}
 \mathcal{M}_j=\left\{m:
 (b_{j,(m+\ell)\bmod12})_{\ell=0}^{4}=\mathbf{o}\right\}.
 \label{eq:marker}
\end{equation}
When the marker has a unique match $m_j$, the information sequence is
\begin{equation}
 \mathbf{q}_j=(b_{j,(m_j+5+\ell)\bmod12})_{\ell=0}^{6}.
 \label{eq:payload}
\end{equation}
The sequence is accepted only if it belongs to the code set in Fig.~\ref{fig5}. Excluding duplicate marker patterns removes coding ambiguity but does not prevent detection errors. A missing or non-unique marker match, or a sequence outside the code set, yields no state report.

Let $\widehat{\mathbf{x}}^{\mathrm{mid}}_j$ be the detected position of the middle illuminated bit of the matched orientation marker. The direction vector $\mathbf{g}_j=(g_{j,u},g_{j,v})^\top$ and heading are
\begin{equation}
 \mathbf{g}_j=\widehat{\mathbf{x}}^{\mathrm{mid}}_j
                  -\widehat{\mathbf{c}}_j,
 \label{eq:direction}
\end{equation}
\begin{equation}
 \widehat{\theta}_j=\frac{180^\circ}{\pi}
       \operatorname{atan2}(g_{j,v},g_{j,u}).
 \label{eq:heading}
\end{equation}
Here $u$ increases rightward and $v$ downward in the image, so positive angles follow the image-coordinate convention from the positive $u$ axis. This heading describes the marker direction. Relating it to a robot body axis or an arena coordinate frame requires the corresponding mounting and calibration conventions.

The output for a successfully decoded observation is position, heading, and information code. A code can represent a sensed color, behavior, or an assigned identifier, depending on the task. A changing state code is not a persistent trajectory identifier. The per-frame procedure does not itself require a temporal prediction model; identity association across frames is outside this formulation.

\subsubsection{Processing Sequence and Computational Cost}
Algorithm~\ref{alg:decoding} summarizes the observation pipeline. It separates the geometric center estimate from the validity conditions for a state and orientation report.
\begin{algorithm}[t]
\caption{Per-frame pose and state observation}\label{alg:decoding}
\KwIn{Image; ring geometry; robot count $n$; codebook}
\KwOut{Valid pose and state observations $\mathcal{R}$}
Detect LED spots and construct $S$ using Eq.~\eqref{eq:spots}\;
Compute $H$ using Eqs.~\eqref{eq:kernel}--\eqref{eq:voting}\;
Select candidate centers using Eq.~\eqref{eq:centers}\;
$\mathcal{R}\leftarrow\varnothing$\;
\ForEach{candidate center $\widehat{\mathbf{c}}_j$}{
 Unwrap the annulus and read $\mathbf{b}_j$\;
 Match the circular marker using Eq.~\eqref{eq:marker}\;
 \If{the marker match is unique}{
  Read $\mathbf{q}_j$ using Eq.~\eqref{eq:payload}\;
  \If{$\mathbf{q}_j$ is in the codebook}{
   Compute $\widehat{\theta}_j$ using Eqs.~\eqref{eq:direction}--\eqref{eq:heading}\;
   Append $(\widehat{\mathbf{c}}_j,\widehat{\theta}_j,\mathbf{q}_j)$ to $\mathcal{R}$\;
  }
 }
}
\Return{$\mathcal{R}$}\;
\end{algorithm}

\begin{figure}[t]
    \centering
    \includegraphics[width=\linewidth]{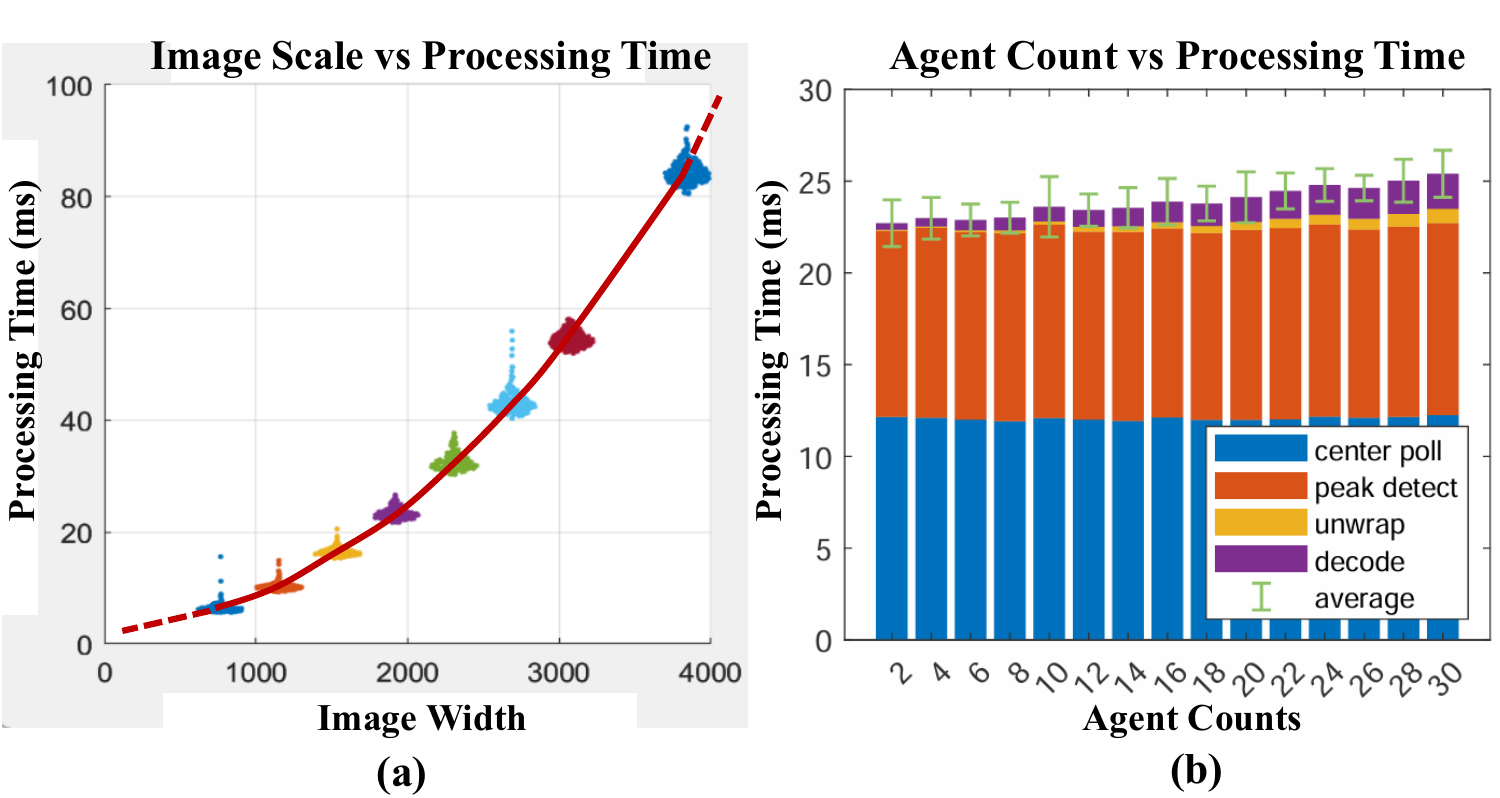}
    \caption{Reported image-processing time. (a) Average processing time versus image width for the tested resolutions, obtained from the camera's native $4096\times2160$ output. (b) Processing-time breakdown versus robot count at $2048\times1280$. Center voting and peak detection account for most of the displayed processing time.}
    \label{fig:6}
\end{figure}

Figure~\ref{fig:6}(a) shows increasing processing time with image width over the tested resolutions. At fixed $2048\times1280$ resolution, Fig.~\ref{fig:6}(b) shows a comparatively modest increase over the displayed range of 2--30 robots. Center voting and peak detection dominate the displayed breakdown, while unwrapping and decoding contribute less. These measurements cover image processing only; they exclude image acquisition, transmission, display updates, and robot response latency.

\subsection{Downstream Interaction}
\subsubsection{Projected Environmental Cues}
Projected light represents environmental information as a spatial intensity or color map. Its values can change according to the experiment rules and observed robot states. Camera observations must be related to the projected arena through calibration. For a planar mapping, let $\widetilde{\mathbf{p}}_c=(u_c,v_c,1)^\top$ and $\widetilde{\mathbf{p}}_p=(u_p,v_p,1)^\top$. The homography relation is
\begin{equation}
 \widetilde{\mathbf{p}}_p\sim
 \mathbf{H}_{cp}\widetilde{\mathbf{p}}_c,\qquad
 \mathbf{H}_{cp}\in\mathbb{R}^{3\times3}.
 \label{eq:homography}
\end{equation}
The symbol $\sim$ denotes equality up to nonzero scale. If $\mathbf{z}=\mathbf{H}_{cp}\widetilde{\mathbf{p}}_c$, the projector coordinates are $(z_1/z_3,z_2/z_3)$ for $z_3\ne0$. This relation applies to the calibrated plane; it does not recover depth or automatically correct offsets between the LED and sensing planes.

\subsubsection{Infrared Commands}
The digital infrared transmitter sends messages containing a recipient identifier, termed a UUID in the system, and a command payload. The protocol uses pulse-position modulation with a 38~kHz carrier. This frequency specifies the carrier, not the payload data rate.

\section{Experiments and Results}\label{sec:experiments}
The evaluation comprises two functional demonstrations and a multi-pheromone case study. Together with the processing-time results, they illustrate the operation of the described implementation. They are not controlled comparisons with RF communication or other localization systems.

\subsection{Path Following and State Reporting}
\begin{figure}[t]
    \centering
    \includegraphics[width=\linewidth]{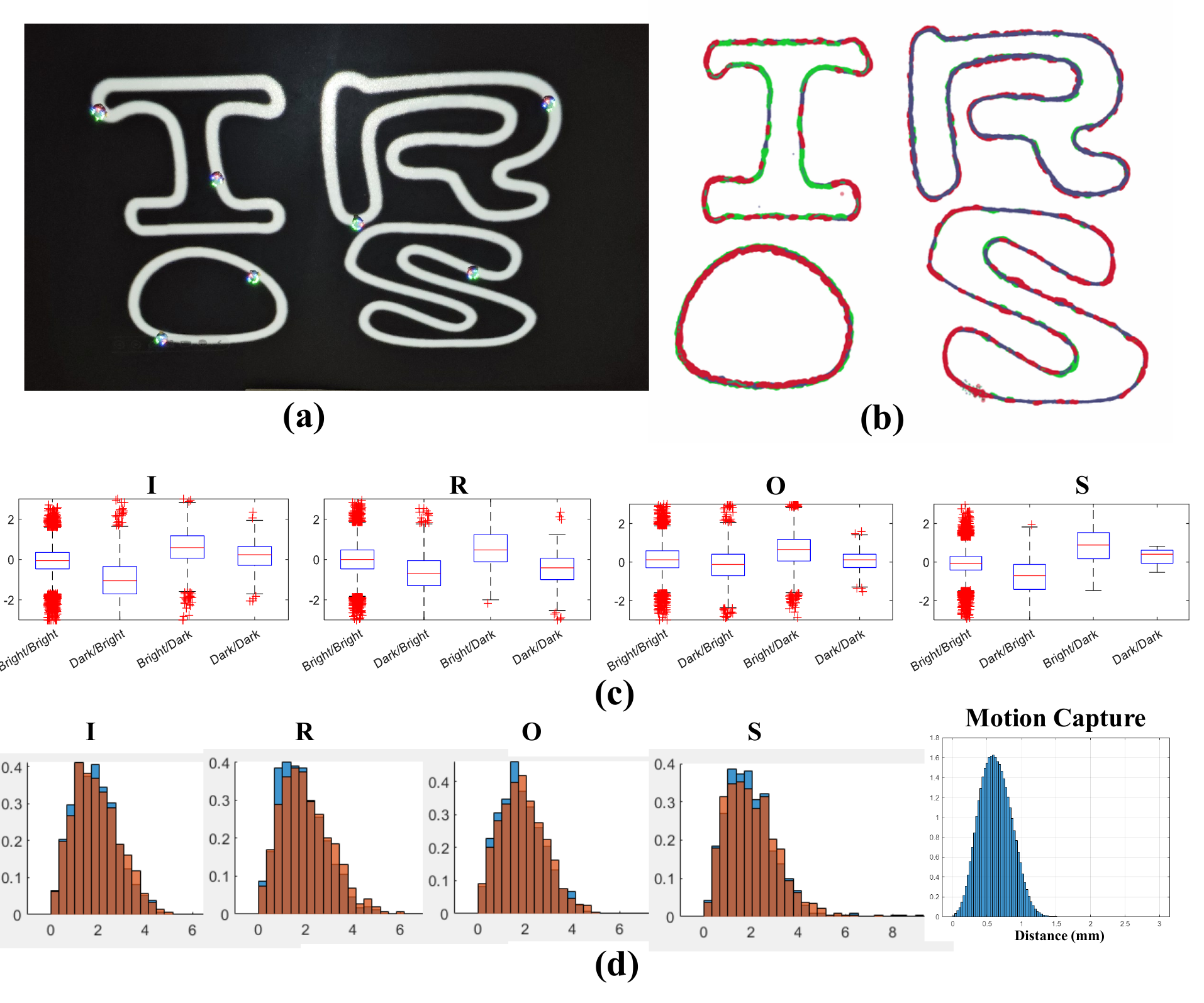}
    \caption{Path-following demonstration. (a) Four projected closed-loop paths. (b) Observed trajectories with reported states: green for right turn, red for left turn, and blue for straight motion. (c) Angular-difference summaries grouped by path and displayed code. The labels Bright--Bright, Dark--Bright, Bright--Dark, and Dark--Dark correspond to straight, left-turn, right-turn, and out-of-range sensing states. (d) Distance-difference distributions and the motion-capture reference. The image scale is approximately 1.22~mm per pixel at $2048\times1280$.}
    \label{fig:7}
\end{figure}

\paragraph{Setup}
Robots follow the closed-loop paths projected in Fig.~\ref{fig:7}(a), using two of the upward-facing color sensors. The LED patterns report the corresponding sensing states: balanced, left-biased, right-biased, or low illumination. A Nokov Mars motion-capture system provides reference observations.

\paragraph{Results and Interpretation}
The trajectories and state annotations in Fig.~\ref{fig:7}(b) illustrate concurrent motion observation and state reporting during path following. Panels (c) and (d) show the reported angular- and distance-difference summaries. However, the available summaries do not specify enough detail about time alignment, coordinate registration, and the definition of each difference to establish absolute position or heading errors against motion capture. The results therefore demonstrate functional operation of the observation and reporting pipeline in this task, not a calibrated accuracy measurement.

\subsection{Color Recognition and Encoding}
\begin{figure}[t]
    \centering
    \includegraphics[width=\linewidth]{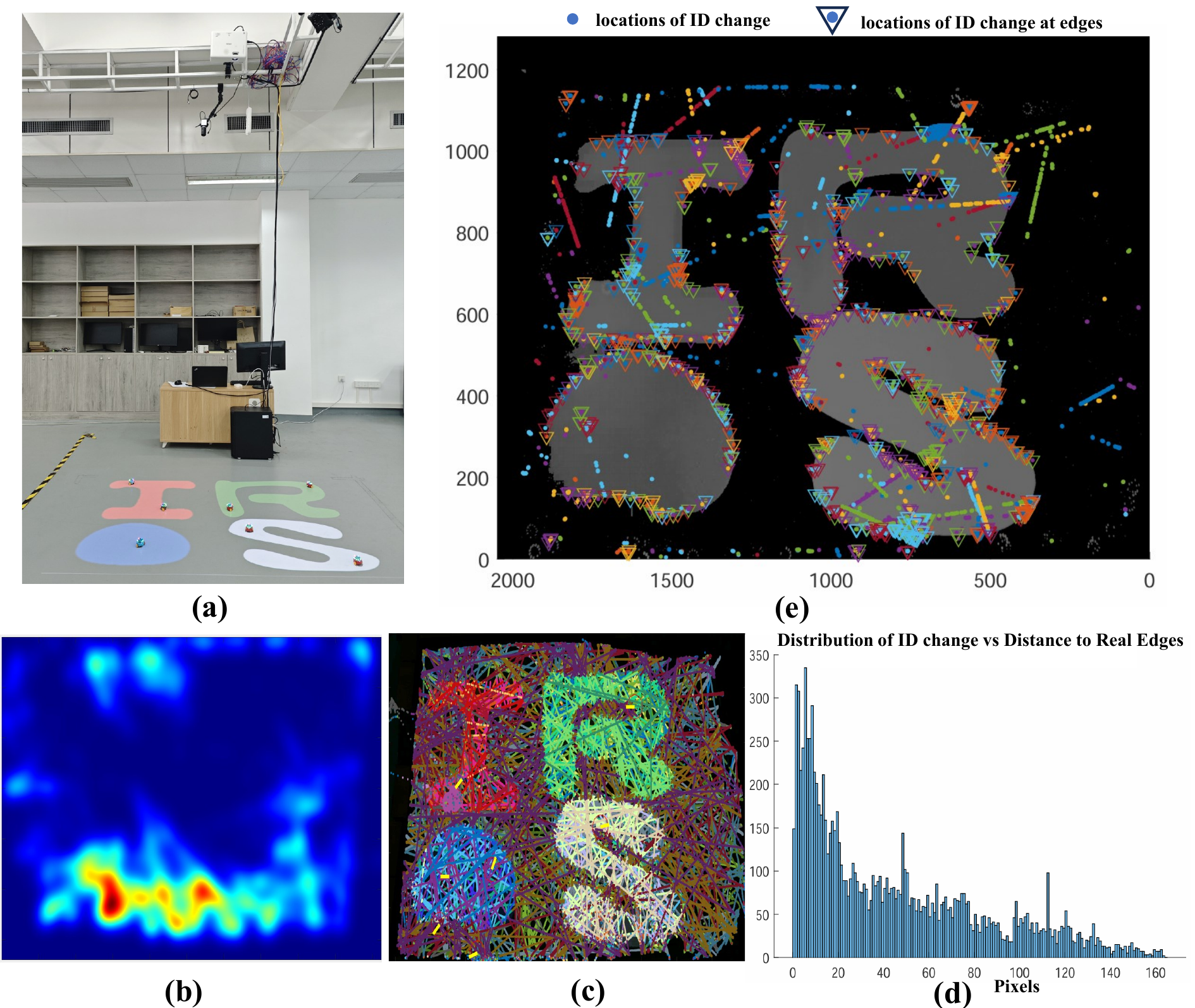}
    \caption{Color-reporting demonstration. (a) Projected color patterns. (b) Spatial distribution of reported color/code discrepancies. (c) Reconstructed color observations. (d) Distances from code-change locations to projected boundaries. (e) Combined map of code changes; triangular annotations mark changes associated with boundaries in the displayed analysis. ``ID change'' in the figure refers to a change in the displayed code, not necessarily robot identity.}
    \label{fig:8}
\end{figure}

\paragraph{Setup}
Robots execute random-walk behavior across the colored patterns in Fig.~\ref{fig:8}(a). Each robot senses the local illumination and represents its detected color state through its LED ring. The overhead camera observes the pattern, and the processor associates the decoded state with an estimated position to reconstruct a color map.

\paragraph{Results and Interpretation}
Figure~\ref{fig:8}(b)--(e) displays the reconstructed observations, discrepancies, and the spatial relationship between code changes and projected boundaries. This task exercises the complete chain from projected illumination and onboard sensing to LED encoding, camera decoding, and spatial registration. Consequently, a mismatch can arise at several stages and cannot be assigned to optical transmission alone. Discrepancies concentrate near the bottom of the map, possibly because of reflections. Code changes near color boundaries are consistent with sensing a transition. The results illustrate spatial color reporting without establishing a bit-error rate or separately measuring sensor and decoder accuracy.

\subsection{Multi-Pheromone Closed-Loop Demonstration}
\begin{figure}[t]
    \centering
    \includegraphics[width=\linewidth]{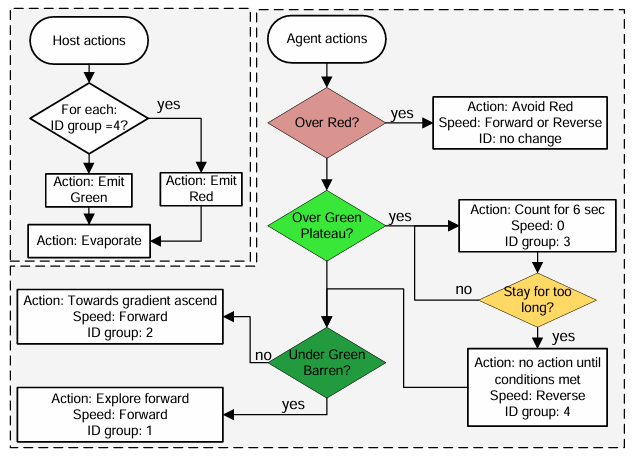}
    \caption{Behavior and host-update flow for the multi-pheromone demonstration. Green cues influence exploration, gradient following, and stopping according to the displayed conditions. Red cues elicit avoidance. The host uses optically reported state groups to determine pheromone deposition and applies evaporation. The figure's ``ID group'' denotes a state-code group.}
    \label{fig:9}
\end{figure}
\begin{figure}[t]
    \centering
    \includegraphics[width=0.8\linewidth]{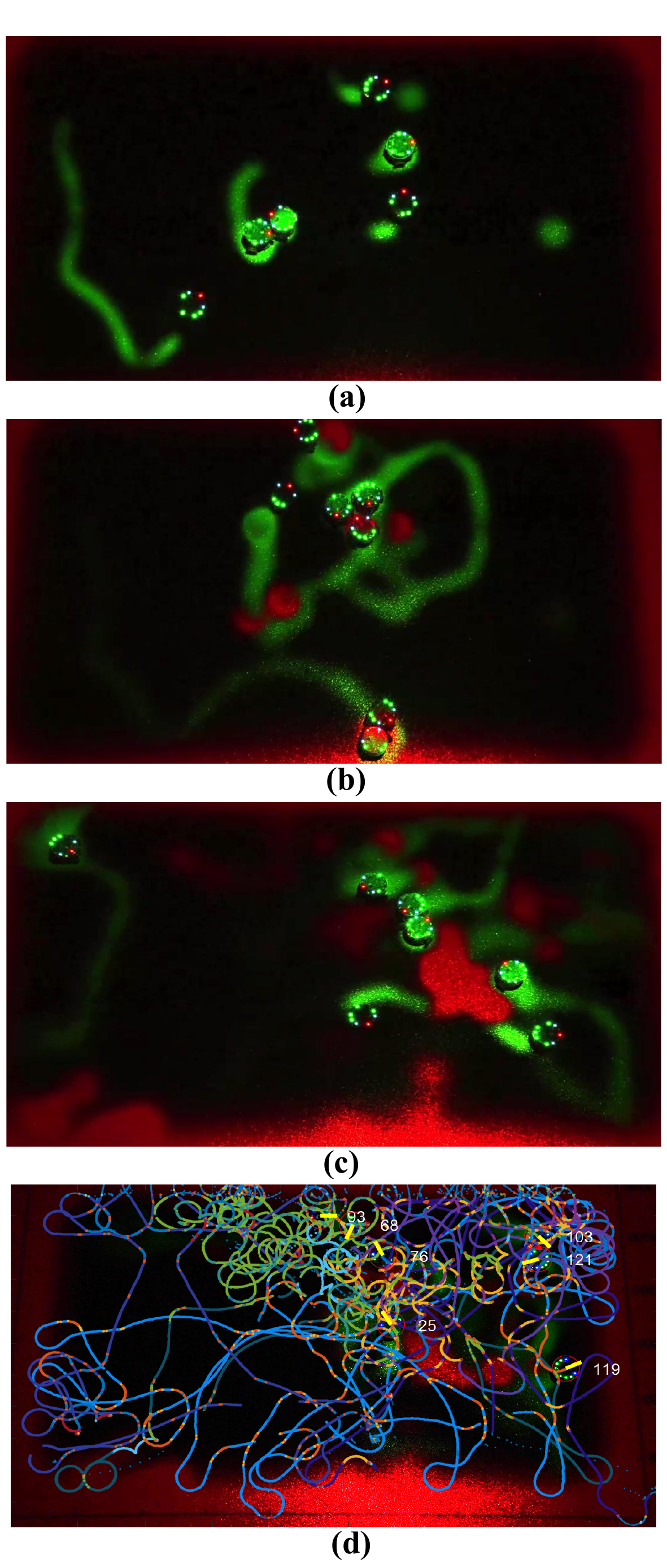}
    \caption{Multi-pheromone demonstration. (a) Initial exploration. (b) A later configuration with aggregation and red cues. (c) Subsequent dispersion as the displayed cue field changes. (d) Sample trajectories.}
    \label{fig:10}
\end{figure}

\paragraph{Setup}
This case study links camera acquisition, pose and state decoding, projected environmental updates, and robot responses in a closed loop. Two virtual pheromones are represented by green and red illumination. The behavior rules are shown in Fig.~\ref{fig:9}. Under weak green illumination, robots explore forward; at intermediate levels they follow the green gradient; at the plateau condition they stop temporarily. Red illumination elicits avoidance. The host associates observed code groups with deposition actions, while evaporation reduces the displayed cues over time. Prolonged residence and the associated state changes lead to red deposition and subsequent avoidance, as represented in the flowchart.

The robot responds to locally sensed light, while the external processor maintains and renders the virtual cue field. This coupling is described at the level of deposition, evaporation, and behavioral transitions; no quantitative field-dynamics model is evaluated here.

\paragraph{Results and Interpretation}
Figure~\ref{fig:10} shows example configurations and trajectories during the demonstration. The sequence illustrates exploration, aggregation, and later dispersion in the evolving cue field. It supports the feasibility of connecting optical state observations to projected feedback in this setup. No control comparison or task-efficiency metric is reported.

\section{Discussion and Conclusions}\label{sec:discussion}
BO-COMM integrates optical pose observation, dynamic state reporting, and downstream interaction for a planar swarm-robotics arena. Its contributions are:
\begin{itemize}
 \item \textbf{Joint pose and state observation.} A ring of LEDs supplies geometric cues for center and heading estimation and a spatial code for changing robot states.
 \item \textbf{Optical channel integration.} Camera-based upstream observation is combined with projected environmental cues and an infrared command pathway on the Colias IV/CEU-PLUS hardware.
 \item \textbf{Functional system demonstrations.} Processing-time plots, path following, color reporting, and a multi-pheromone case study document aspects of the tested implementation and its closed-loop operation.
\end{itemize}

Shared-channel communication contention remains a relevant motivation. By moving selected state reports to spatially distinguishable LED patterns, this architecture reduces the need to transmit those reports through an RF channel. This is a functional alternative, not an experimentally established elimination of congestion. Optical reporting introduces its own constraints: line of sight, image resolution, illumination, and separation of visible features. The finite state-code set also does not remove the need for identity management when a task requires persistent robot identities.

The demonstrated system depends on the specified robot extension, camera geometry, and planar calibration. The measured processing-time trend is limited to the displayed resolutions and robot counts. The experiments do not characterize end-to-end latency, infrared delivery reliability, absolute pose error with a fully specified reference alignment, or robustness across controlled lighting and occlusion conditions. The behavioral examples demonstrate integration rather than comparative swarm-task performance. Within these boundaries, BO-COMM provides a concrete implementation of joint optical pose and state observation connected to a programmable robot environment.

\section*{ACKNOWLEDGMENT}
This work was supported in part by the National Natural Science Foundation of China under Grant 52202492 and Grant 12031003; and in part by the National Undergraduate Innovation Training Program Grant
 S202411078007. 

\end{document}